\def\year{2020}\relax
\documentclass[letterpaper]{article} 
\usepackage{aaai20}  
\usepackage{times}  
\usepackage{helvet} 
\usepackage{courier}  
\usepackage[hyphens]{url}  
\usepackage{graphicx} 
\usepackage{graphicx}  
\usepackage{xcolor}

\usepackage{latexsym}

\usepackage{booktabs} 
\usepackage{algorithm}
\usepackage{algpseudocode}%
\usepackage{multirow}
\usepackage{tabularx}
\usepackage{comment}
\usepackage{graphicx}
\usepackage{amsmath}
\usepackage{tabularx}
\newcolumntype{L}{>{\raggedright\arraybackslash}X}

\title{Towards Quantifying the Distance between Opinions}
\author{
\large{
Saket Gurukar \textsuperscript{1},
Deepak Ajwani\textsuperscript{2}, 
Sourav Dutta\textsuperscript{3,} \thanks{This work was done while the author was at Nokia Bell Labs, Ireland}, 
Juho Lauri\textsuperscript{*}}, \\
\textbf{\large{Srinivasan Parthasarathy\textsuperscript{1}, 
Alessandra Sala\textsuperscript{4}}} \\
\textsuperscript{1} Department of Computer Science and Engineering, The Ohio State University \\
\textsuperscript{2} School of Computer Science, University College Dublin \\
\textsuperscript{3} Huawei Ireland Research Center\\
\textsuperscript{4} Nokia Bell Labs \\
gurukar.1@osu.edu, deepak.ajwani@ucd.ie, srini@cse.ohio-state.edu, \\sourav.dutta2@huawei.com, juho.lauri@gmail.com, alessandra.sala@nokia-bell-labs.com
}

\date{}

\begin{document}
\maketitle

\begin{abstract}

Increasingly, critical decisions in public policy, governance, and business strategy rely on a deeper understanding of the needs and opinions of constituent members (e.g. citizens, shareholders).
While it has become easier to collect a large number of opinions on a topic, there is a necessity for automated tools to help navigate the space of opinions. 
In such contexts understanding and quantifying the similarity between opinions is key.
We find that measures based solely on text similarity or on overall sentiment often fail to effectively capture the distance between opinions. 
Thus, we propose a {\em new distance measure} for capturing the similarity between opinions that leverages the nuanced observation -- {\it similar opinions express similar sentiment polarity on specific relevant entities-of-interest}. 
Specifically, in an unsupervised setting, our distance measure achieves significantly better Adjusted Rand Index scores (up to 56x) and 
Silhouette coefficients (up to 21x) compared to existing approaches. Similarly, in a supervised setting, our opinion distance measure achieves considerably better accuracy (up to 20\% increase) compared to extant approaches that rely on text similarity, stance similarity, and sentiment similarity.
\end{abstract}

\section{Introduction}
\label{sec:introduction}

Crucial decisions in public policy-making as well as in business strategy can be enhanced through a deeper understanding of the diverse opinions and perspectives put forth by relevant stakeholders. While elections, referendums, and market surveys provide important mechanisms for gauging public opinion, in general, they are (i) expensive and (ii) primarily involve selecting from a predefined set of options. Furthermore, many decision processes (i) may not justify the huge expense of referendums, and might also (ii) require a more nuanced analysis of opinions on a more frequent or regular basis. 
This has led to the growth of {\em digital democracy platforms} for continuous collection of public opinions at a significantly less cost, enabling better analysis and alignment of decisions with the viewpoints of the stakeholders. For instance, governing institutions in many democratic countries issue public notices to seek opinions on governing policies (e.g., Net neutrality NPRM issued by the U.S.\ Federal Communications Commission or the Brexit referendum). On a smaller scale, even local city councils call for public consultations on administrative issues such as local property taxes or road works. Similarly, businesses spend a considerable amount of resources to understand and organize customer feedback on products and services. 
In many cases, the collection of opinions is too large to be manually curated. For instance, on the BBC News website, popular articles receive thousands of comments.
Thus, there is a need for automated tools to not only navigate but also assist in understanding the space of all opinions. A fundamental challenge in navigating opinions (or clusters of opinions)  is the need to construct a distance measure that  \emph{quantifies the distance between opinions}. A good distance measure should 
be able to semantically differentiate between opinions that are highly similar 
and opinions that are opposing. Despite the fundamental importance of such an opinion distance function, we note that the traditional approaches are inadequate for this purpose. 

 Many existing approaches for {\em opinion mining} rely on features based on the text-similarity of the opinion documents \cite{mullen2004sentiment} or the overall sentiment orientation of the comment (i.e., whether the overall tone is positive, negative or neutral) \cite{pang2002thumbs}. Unfortunately, such approaches involving text-similarity and sentiment analysis are often inadequate in our problem setting. For instance, consider the diametrically opposite opinions of {\small \tt "In this debate, Hillary looked presidential while Trump came across as manipulative"} and {\small \tt "In this debate, Trump looked presidential while Hillary came across as manipulative"}. Use of text-similarity or overall sentiment based features 
 will categorize the above opinions to be {\bf very similar} as they have the same bag-of-words, similar sentence structure and identical set of sentiment words. In fact, we demonstrate later that text-similarity based features like TF-IDF and also semantic measures like Word-Mover distance and Doc2vec are poor indicators of opinion similarity on many datasets. Furthermore, such features resulted in opinion clusters that have very little agreement with the ground truth clusters, even for cases with only two clusters. 

The above exemplar suggests that to effectively capture the differences between opinions, a measure needs to be able to understand deeper nuances of the semantics of the text. Given the inherent difficulties in capturing the ``true'' opinion distance (as opposed to text-similarity distance),
recent research has focused on a simpler variant of this problem called {\em stance detection} (i.e., whether an opinion is in-favour of, neutral or in-opposition to a target topic). 
However, human opinions are often nuanced and 
organizing them independently into stances can lead to distortion or misinterpretation -- a scenario increasingly crucial in the political arena. For example, on the issue of Brexit, consider the following opinions: (i) ``Brexit will result in economic loss, no doubt, but because it will reduce immigration numbers drastically, it will still be worth it"; and (ii) ``Brexit will result in economic prosperity and huge savings and both the quality and quantity of the immigrants will increase". Both are in favour of Brexit (depicting the same stance), but are still fundamentally different in terms of the opinions as to why the same stance is supported. Existing similarity measures fail to capture such nuanced opinion analysis, which might result in wrong conclusions about what people really opine about a topic.



\begin{figure}[t]
\centering
\includegraphics[scale=0.30]{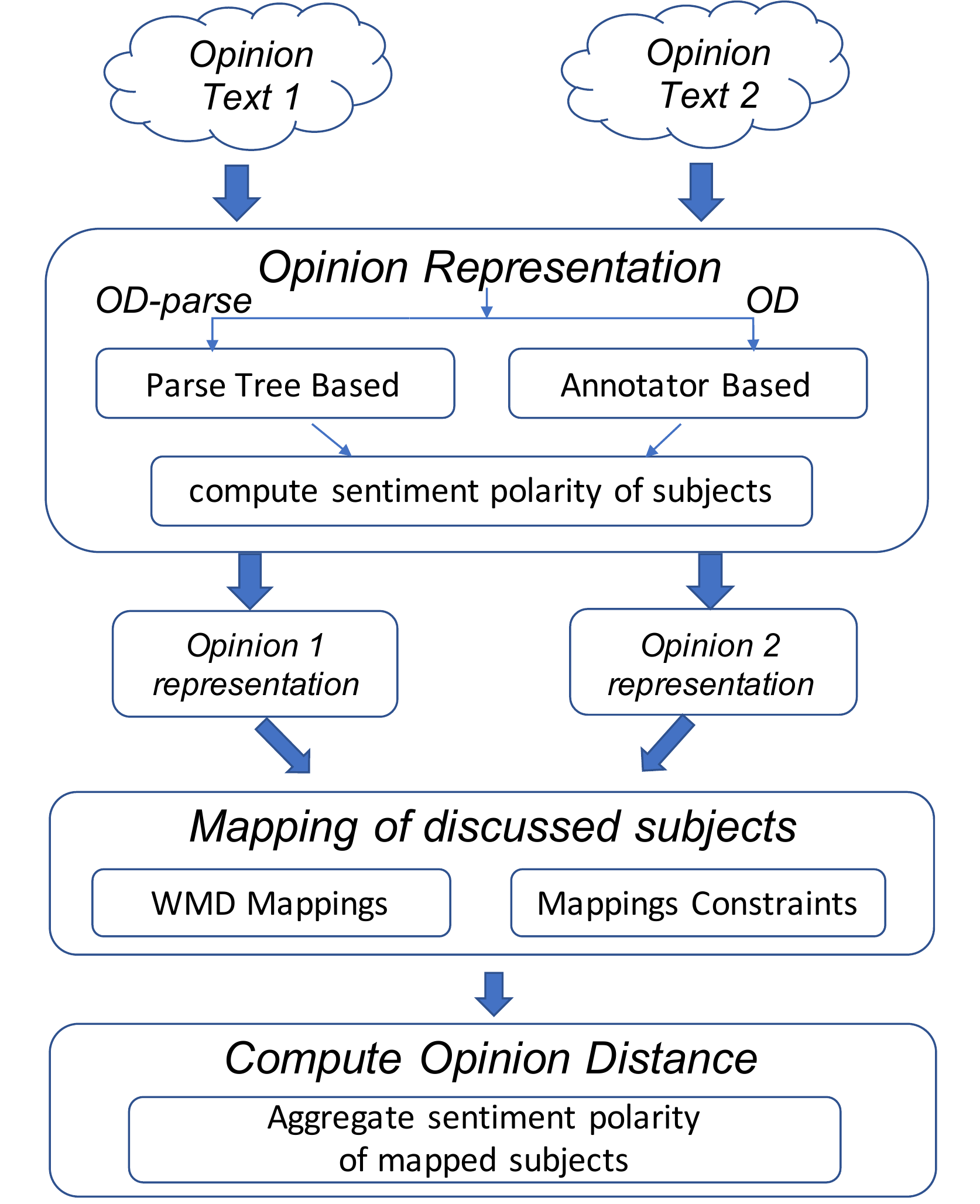}
\caption{Opinion Distance Pipeline}
\label{fig:pipeline}
\end{figure}


\textbf{Contributions:} We address the problem of opinion clustering by proposing a distance measure that  quantifies the semantic similarity between opinions. Leveraging the observation that similar opinions express similar sentiment polarity on the relevant subjects, we represent an opinion in terms of the discussed subjects and the sentiment polarity expressed towards those subjects.\footnote{A similar representation of opinion is considered in \cite{kim2006extracting,kobayashi2004collecting,recupero2015sentilo,liu2005opinion}.} The subjects in the opinion texts are then mapped based on semantic similarity and the opinion distance is defined as an aggregate of sentiment polarity difference towards the mapped subjects. We study a few concrete instantiations of our distance measures and propose a carefully engineered computational pipeline (c.f. Figure~\ref{fig:pipeline}). We also demonstrate improved experimental results (in both supervised and unsupervised settings) on several real-world data-sets along with a use-case study to organize comments on BBC news portal, showcasing the efficacy of the proposed distance measures for organizing opinions over extant approaches on both unsupervised and supervised task settings. 

\section{Background}
%

We begin by briefly reviewing the {\bf key definitions}:
\begin{enumerate}

\item {\it Target Entity}: The subject or topic of interest that frames the discussion or narrative.

\item {\it Aspect}: Characteristics of the target entity.

\item {\it Perspective}: A way of viewing or perceiving the target entity or topic by a person. 

\item {\it Stance}: Subjective disposition towards the topic, typically, characterized as being in favor, neutral towards, or against defined target entity.

\item{\it Opinion}: The statement(s) reflecting and/or justifying the belief or judgment of a person towards a target entity or its aspect(s). Note that {\it two opinions may be very different, even though they may have the same stance towards the target entity}. 

\end{enumerate}

\section{ Related work}
Contrastive opinion modeling~\cite{FSSY12}
relies on computing topic models --  a variant of LDA using Gibbs sampling to estimate the model parameter. The authors subsequently adopt the Jensen-Shannon divergence among the individual topic-opinion distributions to determine contrastive opinions.\footnote{We could not obtain the source code.}
The problem of identifying different perspectives or viewpoints about a topic is addressed by proposing a graph partitioning method which exploits the presence of \textit{social interactions} in order to identify viewpoints \cite{quraishi2018viewpoint}. Our current effort does not rely on social interactions in order to quantify the opinion distance. 
The Author Interaction Topic Viewpoint model (AITV) \cite{trabelsi2018unsupervised} focuses on the task of viewpoint detection of author and post. The AITV topic model levers authors' interactions and encodes the heterophily assumption that difference in viewpoints induces more interactions. However, such interactions are not always available. 

Stance detection has been widely studied with a focus on short-text social media~\cite{mohammad2016semeval} or news articles~\cite{stab2018argumentext,riedel2017simple,awadallah2012harmony}. Many studies ~\cite{riedel2017simple} have shown that features based on n-grams trained with SVM 
are difficult-to-beat baselines for such tasks. 
Similarly, sentiment analysis often rely on dependency parse trees~\cite{socher2013recursive,zhang2018deep} and aspect-based sentiment analysis~\cite{pontiki2016semeval,pavlopoulos2014aspect,titov2008joint} using conditional random field classifiers have also been proposed. However, note that  {\it stance}, {\it sentiment}, and {\it opinion} have nuanced differences as defined earlier.

Sentiment analysis has been used in customer feedback on a brand name. Socher et al. \cite{socher2013recursive} compute the sentiment of a sentence by assigning sentiment to individual words and phrases in the dependency parse tree of the sentence and then recursively aggregate the sentiments in the parse tree to compute the sentiment of the sentence. For more on sentiment analysis, see~\cite{buche2013opinion,pang2008opinion,liu2012survey}. 
Aspect-Based Sentiment Analysis (ABSA)~\cite{pontiki2016semeval,pavlopoulos2014aspect,titov2008joint} is a subfield of sentiment analysis where sentiments towards each aspect are studied. 
However, most of the studies comprises of supervised methods and require the definition of aspects to be known beforehand. 
Note that, in this work, we focus on developing an unsupervised distance measure for opinions. \looseness=-1

There has been considerable work (e.g.,~\cite{kim2006extracting}) on extracting opinion targets and expressions. However, most of the proposed methods are supervised and domain specific~\cite{wiegand2015opinion}.
The Sentilo 
tool~\cite{recupero2015sentilo} identifies the discussed entities in opinion and the sentiment expressed towards the entity in an unsupervised and domain-independent way. However, the methods proposed in opinion target and opinion expression extraction literature cannot be extended straightforwardly to compute opinion distance. For instance, the issue of opinion subject polysemy while computing opinion distance is nontrivial to solve.


We stress that the problems addressed in the above studies are different than ours. To our knowledge, there is no work on an ``opinion distance measure" focusing on quantifying the similarity or dissimilarity between different opinions.

\section{Distance measures}
\label{sec:distance}

For designing a distance measure for opinions, we leverage the observation that similar opinions express similar sentiment polarity on the relevant discussed subjects. We propose a set of measures based on aggregating the difference in sentiment polarity of words associated with the subjects. 

The key to designing a good distance measure lies in the representation of the opinion itself. That is, instead of merely relying on the overall tone/sentiment of the opinion or just considering the bag-of-words or collection of n-grams representations, we propose a more nuanced representation of opinion in terms of the discussed subjects and the sentiment polarity expressed towards those subjects. To this end, our distance computation framework: (i) extracts the opinion subjects discussed in the text, (ii) identifies the words associated with the subjects that express opinions towards those subjects, and (iii) computes the sentiment polarity of the associated words. 

Given two opinions ($O_1$ and $O_2$), each represented in terms of a tuple of opinion subjects and sentiment polarity towards the corresponding subject within each opinion, their distance is computed by: (i) first, mapping opinion subjects in $O_1$ to corresponding subjects in $O_2$ (and vice versa) based on their semantic similarities; and (ii) second, aggregating the difference of sentiment polarities expressed on the correspondingly mapped subjects across both opinions. Figure~\ref{fig:pipeline} presents the different steps in our pipeline. Note that it is crucial to have a reasonably accurate semantic mapping between subjects, as common subjects might be referred to with different phrases (or surface-forms) in different opinions.

\textbf{Mapping opinion subjects:} To map the different subjects among opinions, we first create a bipartite graph by computing a semantic similarity score between the subjects discussed in the opinions. This graph is then used to compute the mapping between the opinion subjects using Word Mover Distance (WMD)~\cite{kusner2015word}, which aims at computing a minimum weight \emph{perfect} matching between the opinion subjects.
A detailed instantiation of this step in our framework is presented later section. 

\textbf{Aggregating polarity difference:} Let $S_{i}^{1}$ and $S_{j}^{2}$ be the $i$\textsuperscript{th} and $j$\textsuperscript{th} opinion subjects in opinions $O_1$ and $O_2$, respectively. Also, let $\text{pol}(S_{i}^{k})$ represent the expressed polarity towards opinion subject $i$ in opinion k ($O_k$). Then, the opinion distance between $O_1$ and $O_2$ is


\begin{equation}
\label{distance_equation}
OD (O_1, O_2) \; = \; \frac{\sum\limits_{(i,j) \in \cal{M}}  \, \textit{f}\; (\; pol(S_{i}^{1}) \; , \; pol(S_{j}^{2}) \;)}{2 \times | \cal{M} |} 
\end{equation}
where $\cal{M}$ is the set of mapped opinion subjects. $i$ and $j$ are opinion subjects in $O_1$ and $O_2$, respectively and \textit{f} is the difference function defined as 

\[
\resizebox{1.0\hsize}{!}{
$
    \textit{f}(x, y) = 
\begin{cases}
    |x - y| ,& \text{if  x, y are absolute values}\\
    JSD(x,y) \ or \ EMD(x,y), & \text{if  x, y are distributions}
\end{cases}
$
}
\]
Here, JSD denotes Jenson-Shannon divergence and EMD denotes Earth Mover Distance. Both JSD and EMD are symmetric measures and EMD does not suffer from arbitrary quantization problems \cite{rubner2000earth}. Note, that the value of $OD(O_1,O_2)$ lies between [0,1]. \\

 \textbf{Example:}
Consider an example with two opinions having opinion subjects as shown in bold below:\\  \textit{$O_1$: ``\textbf{Video games} increases violent tendencies among \textbf{youth}."} \\ 
\textit{$O_2$: ``\textbf{Researchers} have confirmed the potential positive effects of \textbf{computer games} and \textbf{media contents.}"} \\

We first compute the representation of the two opinions in terms of opinion subjects and polarity. So, $O_1$ is represented by the vector [(``Video game", -1), (``youth", 0)] and $O_2$ is represented by [(``researchers", 0), (``computer games", +1), (``media contents", +1)]. We then compute the semantic distance matrix which contains the semantic distance between the subjects in $O_1$ and $O_2$. Based on this matrix, our framework identifies that the subjects ``video game" and ``computer games" are highly similar, while the others are not. Let $S_{i}^{1}$ be the opinion subject ``Video game", while $S_{j}^{2}$ be the opinion subject ``Computer game". Further, the polarity of $S_i^1$, $\text{pol}(S_{i}^{1})$ is computed as ``-1", while $\text{pol}(S_{j}^{2})$ is ``+1"; and the difference function \textit{f}($\text{pol}(S_{i}^{1})$, $\text{pol}(S_{j}^{2})$) is $2$. Since this is the only mapping subject pair ($|\cal{M}|$=1), the final opinion distance $\text{OD}(O_1, O_2)$ is obtained as $1$ (max. value). Thus, the distance  is large.

\section{Opinion representations} 

In this section, we present two alternative representations for identifying opinion subjects. The first representation considers noun-phrases as opinion subjects and the dependent adjectives, adverbs, and verbs as associated words. The second representation consists of disambiguated concepts as opinion subjects with the words surrounding the subject as associated words. While the former relies on carefully defined rules to identify noun phrases and a careful analysis of the dependency parse tree to compute the polarity of the dependent associated words, 
the latter uses the weighted aggregation of the polarity of the neighboring words. 
We note that the latter is more efficient and avoids the computationally expensive and error-prone step of dependency parsing. We refer to the distance measures computed using the above variants as {\em OD-parse} and {\em OD}, respectively, and empirically compare their effectiveness in the experiments section. 

\subsection{Noun-phrase Representation}
\label{sec:opinion_parse_tree}
A parse tree represents the structure of a text string based on the syntax of the input language (in our case, the English language). A dependency parse tree captures the dependencies between different linguistic units in a sentence. We use Stanford CoreNLP~\cite{manning2014stanford} for part-of-speech (POS) tagging, coreference resolution and dependency parsing. Note, our framework is agnostic to such choices, and other tools like Google SyntaxNet~\cite{AAWSPGPC16} could also be used.

\textbf{Opinion subjects extraction:} In this approach, we consider all noun phrases in the opinion post as opinion subjects. We define a noun phrase based on some carefully defined rules and named-entities like Person, Organization, and Location. 
An exemplar of such a rule is, \\

\noindent{
\scriptsize
$\left(<\text{NN}.*><\text{POS}>?\right)+<\left(\text{OF}|\text{THE}|\text{IN}\right)>?\left(<\text{NN}.*><\text{POS}>?)\right)+$
} \\

 where $\text{NN}$ denotes a noun word (based on Stanford CoreNLP POS tagging), $\text{POS}$ denotes a possessive form, and the symbols `.*', `?', `+' are regex symbols as defined in Stanford CoreNLP. 
Observe, the above rule can capture even complex noun phrases like ``J.\ R.\ R.\ Tolkien's Lord of the Rings". \looseness=-1

\begin{figure}[t]
\centering
\includegraphics[width=0.85\linewidth]{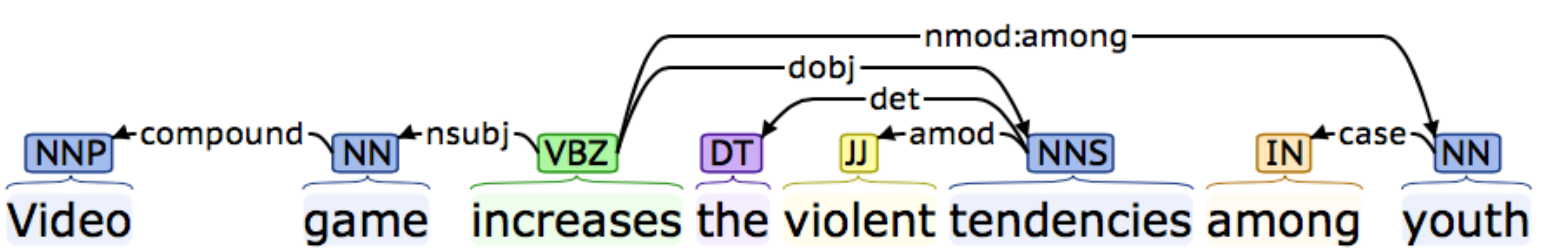}
\caption{Dependency tree of sample sentence}
\label{fig:sampledepen}
\end{figure}

\textbf{Opinion expression extraction:} To calculate the sentiment polarity expressed on the subject, we first identify all the related verbs, adverbs, and adjectives dependent on the noun phrase. For this, we use co-reference resolution 
and dependency parsing. For instance, Figure~\ref{fig:sampledepen} shows the dependency parse tree of opinion $O_1$: \textit{``Video game increases the violent tendencies among youth."}.

 Here, the noun phrase ``Video game" would be extracted as an opinion subject while the verb ``increases" is related and is part of how the subject ``video game" is expressed. However, an efficient representation of opinions should also involve the terms ``violent tendencies" in the opinion expression of ``Video game". To extract the set of words for opinion expressions, we carefully defined $14$ rules using Stanford CoreNLP's Semgrex pattern matching system (the rules are shared in the appendix). 

\textbf{Polarity of opinion subjects:} We use the IBM Debator~\cite{toledo2018learning} sentiment lexicon to get the sentiment polarity score associated with individual words.\footnote{Note that if an individual word has a negation modifier present in the dependency tree, then we multiply the sentiment polarity of that word by -1.}  We aggregate the sentiment polarity scores of different words in the opinion expression in order to compute the sentiment expressed towards the corresponding subject. For this, we consider two techniques:

 1. {\it Average and then discretize the polarity scores.} If the average sentiment polarity score of the words in the opinion expression of an opinion subject is negative, we assign a -1 score to the opinion subject, otherwise we assign it the score +1. 

 2. {\it Consider entire distribution of polarity scores.}


Overall, we found the parse-tree based approach to be computationally intensive. Furthermore, simply treating all noun phrases as opinion subjects resulted in the misidentification of many opinion subjects. For instance, in our example, ``tendencies" would also act as an opinion subject. Another issue is that we do not capture any semantics, so synonyms get identified as different subjects while polysemies get identified as the same subject. Also, efficient extraction of opinion expressions requires a large number of rules and identifying all such rules is a manual and cumbersome process.

\subsection{Disambiguated concept representation}
\label{opinion_light_ref}

In order to resolve the issues with the previous approach, we now describe an alternative approach.
%

\textbf{Opinion subjects extraction:} Ideally, the opinion subjects should be the key entities and concepts discussed in the opinion post. They should have a canonical representation, independent of the exact noun phrase used to discuss them, which can be achieved by leveraging {\em named-entity disambiguation} approaches. 
In this work, we use the popular TagMe spotter to identify relevant noun phrases and the TagMe API~\cite{ferragina2010tagme} for disambiguating the noun phrases to Wikipedia pages. The disambiguation process aids in understanding the noun phrases referring to the same entity/concept as well as those that are referring to entities/concepts that are semantically similar. For example, TagMe maps both the phrases ``Video game" and ``Electronic game" to the Wikipedia page of ``Video game", while it maps ``Computer game" to the semantically similar Wikipedia page titled ``PC game".


\textbf{Opinion expression extraction:} To avoid the error-prone and computationally intensive process of fine-grained dependency parsing, we use all the adjacent sentiment words of an opinion subject $S$ as the opinion expression towards $S$, similar to~\cite{bar2017stance}. However, not all adjacent words are treated equally. The influence of an opinion expression towards an opinion subject is weighted based on how far the opinion expression is from opinion subject in the sentence. Although in this relaxation many unrelated words may end up influencing the computation of sentiment polarity towards a subject, we show in the experiments section that this approach performs well in practice.

\textbf{Polarity of opinion subjects:} Similar to the previous approach, we use the IBM Debator~\cite{toledo2018learning} sentiment lexicon to find the sentiment polarity of words in an opinion expression. The sentiments of different words in the opinion expression are then aggregated by either considering the entire distribution or taking the average and then discretizing it. However in this case, before taking the average, we reweigh the sentiment value of a word by penalizing the sentiment score by a factor of $\sqrt{d}$ where $d$ is the token distance of the word from opinion subject in the sentence. This ensures that the further a word is from the opinion subject, the less influence it has on its polarity score. Furthermore, 
if a polarity shifter (negator words like ``no", ``not" and ``cannot") is present in the opinion expression, we reverse the polarity of the word. The complete list of polarity shifter words considered by our framework is reported in Table \ref{polarity_shifters}.

\begin{table}
\resizebox{0.95\linewidth}{!}{
\begin{tabular}{p{\linewidth}}
\toprule
\textbf{List of sentiment polarity shifters} \\ \toprule
\small{
`no', `not', `negation', `none', ``n't", `inconclusive', `without', `excluded', ``incompatible", ``prevent", `exacerbate',   `reduce',  `less', `rarely',  `displaced', `relocation', `dislocation', `higher than', `relocate', `resettled', `re-housed', `cannot', `limit', `outweigh', `unless', `little act', `get even'} \\ 
\toprule
\end{tabular}
}

\caption{Sentiment polarity shifters used across all datasets. \looseness=-1}
\label{polarity_shifters}
\end{table}

We then take the weighted average of distance-weighted sentiment scores (after polarity shifters). Hence, if $p$ and $n$ refer to the weighted sum of positive and negative sentiments associated with the subject, the weighted average is given by $(p-n)/(p+n+1)$. This is similar to~\cite{feldman2011stock} and we empirically show that such a representation of opinion expression results in identifying better opinion clusters.
Note that if an opinion subject is present in a sentence, all the words present in that sentence would influence the  weighted computation of sentiment score of the subject. For example, our opinion subject ``Video game" would now be represented as average weighted polarities of ``increased", ``violent", ``tendencies", and ``youth".

Interestingly, one could consider including the polarity of the opinion subject itself for computing the average polarity. However, this might incorrectly represent the opinion. For instance, consider the opinion ``Genocide is good''. Here, the opinion subject ``Genocide" will have a positive polarity,  correctly capturing what is expressed. Including the polarity of the subject, would lead to ``Genocide" having a neutral or negative polarity -- different from the expressed opinion.

\section{Mappings between opinion subjects}
\label{wmd_mappings}


To map the opinion subjects in the different opinions we create a bipartite graph by computing the semantic similarity score between the opinion subjects. 
Semantic similarity can be computed using embedding techniques such as word2vec~\cite{mikolov2013efficient} or doc2vec~\cite{le2014distributed}. In addition, the latter can use text similarity between corresponding Wikipedia page abstracts, number of common in-links and out-links between their Wikipedia pages, and so on. We then use the bipartite graph to compute the mapping between the opinion subjects using Word Mover Distance (WMD)~\cite{kusner2015word}. An efficient linear time implementation of WMD exists~\cite{pele2008} to compute ``the minimum distance that the embedded words of one document need to travel to reach the embedded words of another document". The underlying flow matrix in WMD computation provides a mapping between the subjects among the opinions.

 WMD aims at computing a minimum weight \emph{perfect} matching between the subjects. However, in practice, there may not be a good semantic mapping between all the subjects in the opinions, even if they are on the same topic.
Thus, the WMD mapping may also include subject pairs that have high semantic distance, resulting in erroneous comparisons. If two dissimilar subjects get mapped in WMD computation the overall distance between two increases. To address such erroneous mappings, we remove all pairs from the mapping whose semantic distance is greater than a user-defined threshold. 
If no mappings between two opinions exist after the semantic distance threshold, we set their opinion distance as undefined.

\section{Experiments}
\label{sec:experiments}

\subsection{Datasets and Empirical Setup} 

{\bf Datasets:}  One issue in the validation our proposed opinion distance 
is the lack of publicly available annotated ground truth datasets. However, there are many good benchmarks available for a restricted version of opinion clustering, namely stance detection. In stance detection, there are generally two labels -- one supporting the topic-of-interest and the other opposing it. 
The assumption we make for evaluating opinion distances on the stance datasets is that ``similar claims demonstrate similar stances (opinions)". 
We collect the stance detection datasets from IBM Debator Claims~\cite{bar2017stance} and Arguments~\cite{hou2017argument} projects. Table \ref{tab:dataset} provides a complete list of 18 datasets used in our evaluation.  We also evaluate our proposed methods against existing stance baselines~\cite{pontiki2016semeval}. 

Additionally, we curate a nuanced opinion dataset from the \url{civiq.eu} platform discussing whether the Seanad (upper house of parliament in the Republic of Ireland) should be abolished. This dataset has three expert-curated opinion clusters~\cite{civiq_expert}\footnote{We plan to release our annotations and dataset publicly.}: 
(1) Abolish the Seanad, (2)  Reform the Seanad instead of abolishing it, and (3) Seanad is ineffective but keep it until Dail is reformed to save the democracy. Note that the stance of the last opinion cluster is aligned with the second cluster -- both against abolishing the Seanad. However, the general opinion expressed in the last cluster is to have the institution of Seanad only till the lower house of the parliament, Dail, is appropriately reformed. 
This is a nuanced argument not easily captured by existing methods.
 


\begin{table}[t]
\centering
\resizebox{0.95\linewidth}{!}{
\begin{tabular}{p{\linewidth}}
\toprule
\textbf{ Datasets} \\
\midrule

Seanad Abolition  (25), Pornography  (52), Gambling  (60), Video Games  (72), National Service   (33), Monarchy   (61), Hydroelectric Dams   (110), Keystone pipeline   (18), Democratization   (76), Open-source Software   (48), Intellectual Property    (66), Atheism   (116), Education Voucher Scheme   (30), One-child policy China   (67), Austerity Measures   (20), Affirmative Action   (81), Housing   (30), Trades Unions   (19) \\
\bottomrule
\end{tabular} 
}
\caption{Dataset information. The number in brackets represents the number of opinions.}
\label{tab:dataset}
\end{table}

 %



\section{Methods}
\label{baselines}
In this section, we explain the selected baselines and the parameter setting for benchmarking the performance of the competing approaches. 

 \textbf{TF-IDF:} The distance between two opinions is computed as the cosine distance between their term-frequency inverse-document frequency (TF-IDF) vectors of those opinions~\cite{schutze2008introduction}. Following standard practice, we remove the stop words while computing the TF-IDF. 

 \textbf{ WMD:} The distance between two text documents can be captured using the recently proposed Word Mover Distance (WMD)~\cite{kusner2015word} -- which in our setting would enable the capture of the distance between opinions. 
We use a pre-trained word2vec embedding~\cite{mikolov2013efficient} trained on GoogleNews Corpus for computing the WMD. 


 \textbf{Sent2vec:} The distance between two opinions 
is the cosine distance between Sent2vec~\cite{pgj2017unsup} embeddings of those opinions. We lever the pre-train sent2vec-wiki-unigrams model 
for computing sentence embeddings. 

 \textbf{Doc2vec:} The distance between two opinions 
is the cosine distance between Doc2vec~\cite{le2014distributed} embeddings of those opinions. We pre-train the Doc2vec methodology on Wikipedia articles. 

BERT: The distance between two opinions is the cosine distance between BERT ~\cite{devlin-etal-2019-bert} embeddings of those opinions. The embedding of opinions is computed by averaging the tokens' BERT embeddings using bert-as-a-service tool ~\cite{xiao2018bertservice}.

Our proposed methods are \textbf{OD-parse}  and \textbf{OD}. Unless noted otherwise, we use Doc2vec embeddings trained on Wikipedia articles with cosine distance for computing the semantic distance. The semantic distance threshold for our framework is set to 0.3.
For TagMe, we select a link probability threshold of 3\%.


\subsection{Experimental results} 

We evaluate the methods in two different settings.

 $\bullet$ \textbf{Unsupervised Setting}: We examine the utility of the proposed opinion distance measure in an unsupervised setting for opinion clustering.
For the competing baselines, we compute the all-pair opinion distance matrix, and then evaluate the distance measure in this unsupervised setting based on clustering quality and on the intra- and inter- cluster distances. The clustering quality is measured using the Adjusted Rand Index (ARI), while evaluations based on intra and inter clusters distance is done with the Silhouette coefficient. 
We set the number of clusters equal to the number of unique opinion labels present in the dataset. 
We perform k-means clustering and spectral clustering and report the best ARI for all approaches. 
The quality of the identified opinion clusters (a snapshot) is shown in Table~\ref{table:cluster_result}. We observe that:

\begin{table}[t]
\small


\label{table:cluster_quality}

\resizebox{0.95\linewidth}{!}{
\def\arraystretch{1.0}
\begin{tabular}{@{}lrrrcrrrcrrr@{}}
\toprule
 & \multicolumn{2}{c}{\begin{tabular}[c]{@{}c@{}}Seanad \\ Abolition \end{tabular}} & & \multicolumn{2}{c}{\begin{tabular}[c]{@{}c@{}}Video \\ Games \end{tabular}} & & \multicolumn{2}{c}{Pornography}\\
\cmidrule{2-3} 
\cmidrule{5-6} 
\cmidrule{8-9} 
Methods & ARI   & $Sil$  && ARI & $Sil$  &&  ARI & $Sil$ \\ 
 
\midrule

TF-IDF  & 0.23 & 0.02 && -0.01 & 0.01 && -0.02 & 0.01 \\
WMD  & 0.09 & 0.01 && 0.01 & 0.01 && -0.02 & 0.01  \\
Sent2vec & -0.01 & -0.01 &&  0.11 & 0.06 && 0.01  & 0.02  \\
Doc2vec  & -0.01 & -0.03 && -0.01 & 0.01 && 0.02 & -0.01  \\
BERT  & 0.03 & -0.04 && 0.08 & 0.05 && -0.01 &  0.03 \\
OD-parse & 0.01 & -0.04 && -0.01 & 0.02 && 0.07  & 0.05  \\
OD & \textbf{0.54} & \textbf{0.31}  && \textbf{0.56} & \textbf{0.42} && \textbf{0.41} & \textbf{0.41} \\

\bottomrule
\end{tabular}

}

\caption{ARI and Silhouette coefficient scores.} 
\label{table:cluster_result}
\end{table}



 \textbf{The opinion distance measure captures the nuances among opinions:}  We see that {\em OD} significantly outperforms the baseline methods and the {\em OD-parse} variant\footnote{{\em OD-parse} fails to perform efficiently possibly due to poor opinion representation induced by erroneous opinion expression extraction based on the dependency parse tree.}. The selected datasets contains set of contrasting opinions with same bag-of-words and as hypothesized earlier the existing text-similarity methods cannot understand the nuances among the opinions -- visible from the low ARI and Sil scores. On the other hand, $OD$ achieves high ARI and Sil scores, and seems to capture the nuances among opinions. Additionally, the noticeably high ARI and Silhouette coefficient values for OD seems to validate the observation of ``similar opinions express similar sentiment polarity on discussed subjects".

 \textbf{Existing distance measures fail to capture nuances among opinions:}
From the above table, we observe that the text-similarity based baselines, such as TF-IDF, WMD and Doc2vec achieving ARI and Silhouette coefficient scores of close to zero on the ``Video Games" and ``Pornography" datasets (barely providing a performance improvement over random clustering, i.e., a zero ARI score). 
A possible source for the poor performance might be the underlying assumption that similar opinions should have common words and/or a similar word distribution -- which does not hold in cases of opinion as depicted in previous sections. A notable exception is the ``Seanad Abolition" dataset, where TF-IDF performs relatively better than WMD, Sent2vec and Doc2vec. Drilling down we find that a highly discriminatory word ``democracy'' occurs in only one of the clusters (namely ``Save the Democracy") - explaining its performance on this dataset.





\begin{table}[t]
\centering
\scriptsize

\resizebox{0.85\linewidth}{!}{

\begin{tabular}{@{}lcccc@{}}
\toprule
{Baselines} & \begin{tabular}[c]{@{}c@{}}  Seanad \\ Abolition \end{tabular} &  
 \begin{tabular}[c]{@{}c@{}} Video \\ Games \end{tabular}  & 
Pornography \\

\toprule


Unigrams & 0.54 & 0.66  & 0.63 \\
Bigrams & 0.54 & 0.64  & 0.56 \\
LSA & 0.68 & 0.57 & 0.57 \\
Sentiment & 0.35 & 0.60  & 0.69 \\
Bigrams & & & \\
\quad + Sentiment& 0.43 & 0.58 & 0.66 \\
\toprule
TF-IDF & 0.50 & 0.65  & 0.57 \\
WMD & 0.40 & 0.73  & 0.57 \\
Sent2vec & 0.39 & 0.79  & 0.70 \\
Doc2vec & 0.27 & 0.51 & 0.56 \\
BERT  & 0.46 & 0.84  & 0.68 \\ 
\toprule
Unigrams  &  &  &  \\ 
\quad + Bigrams  & 0.40  & 0.64 & 0.78 \\ 
\quad + Sentiment  & 0.24 & 0.54  & 0.54 \\ 
\quad + LSA  & 0.73 & 0.51  & 0.58 \\ 
\quad + TF-IDF  & 0.42 & 0.65  & 0.56 \\ 
\quad + WMD  & 0.48 & 0.73  & 0.53 \\ 
\quad + Sent2vec  & 0.56 & 0.59  & 0.66 \\
\quad + Doc2vec  & 0.31 & 0.56  & 0.47 \\ 

\toprule
OD-parse & 0.50 & 0.58   & 0.53 \\ 
OD & 0.71 & \textbf{\underline{0.88}}   & 0.88 \\ 
\toprule
OD  &  &  &  \\ 
\quad + Unigrams & 0.83 & {\it \underline{0.86}}  & {\it 0.88 }\\ 
\quad + Bigrams & \textbf{\underline{0.87}} & 0.85 & {\it \underline{0.88} }\\ 
\quad + Sentiment  & 0.64 & {\it \underline{0.86}}  & 0.86 \\ 
\quad + LSA  & {\it\underline{0.84}} & 0.82  & \textbf{\underline{0.90}} \\ 
\quad + WMD  & 0.75 & 0.82  & 0.86 \\ 
\toprule

\end{tabular}
}
\caption{The quality of opinion distance when leveraged as a feature for multi-class classification. Each entry in + \textit{X} feature should be treated independently. The second best result is italicized and underlined.}
\label{table:cluster_quality_supervised}
\end{table}

 $\bullet$ \textbf{Supervised Setting}: We now examine the effectiveness of our approach in a supervised setting. 
Here, we leverage the idea that distance or similarity based problems can be reformulated as standard classification problems by treating pairwise similarities as features to a downstream classification method like SVM~\cite{GraepelNIPS98,Pekalska01,Noble03}. 
Specifically, for all the approaches, we treat the distance measure as a feature by considering each row $i$ of the distance matrix as a feature vector for opinion $i$ and the task is to check whether two opinions have same label or not.
For all the classification experiments, unless otherwise noted, we use SVM as classifier with RBF kernel.\footnote{Note that, we achieve similar score for the distance measure with Logistic Regression classifier.}
We perform hyperparameter tuning using both grid search (with 5-fold cross validation)  and auto-sklearn~\cite{NIPS2015_5872} (with default `holdout' strategy). We use two tuning strategies because we face the problem of overfitting for smaller datasets using only one of the tuning strategy. Since we use two tuning strategies, the hyperparameter tuning is done on train split (70\%) and we report the best results on test split (30\%) averaged over 3 runs. For evaluation, we rely on the average weighted F1 measure for classification accuracy.  
For completeness, here we also compare 
against unigram or n-gram based classifiers that typically work well in such settings~\cite{pontiki2016semeval,somasundaran2010recognizing}. 
The classification performance of the baselines is reported in Table~\ref{table:cluster_quality_supervised}. We observe that:

\textbf{SVM with only OD features outperforms many baselines:}
We see that on ``Video Games" and ``Pornography" datasets, the classification performance based on SVM with only OD is significantly better than the SVM with any other combination of features excluding OD. For the ``Seanad Abolition" dataset, there is one exception: SVM with unigrams and LSA features performs slightly better than OD. As discussed earlier, this can be attributed to the discriminating word ``democracy".  

 \textbf{SVM with OD and baseline features further improves classification performance:} We see that SVM with OD and bigrams achieves the best multi-class classification performance on the ``Seanad Abolition" dataset. On the ``Pornography" dataset, we observe SVM with OD + LSA to improve classification performance by nearly 2\%.





\begin{table}[t]
\centering
\resizebox{0.95\linewidth}{!}{
\begin{tabular}{@{}llrrrcrrrcrrr@{}}
\toprule
& \multicolumn{1}{l}{\begin{tabular}[l]{@{}l@{}}Difference \\ Function \end{tabular}}  & \multicolumn{1}{c}{\begin{tabular}[c]{@{}c@{}}Seanad \\ Abolition \end{tabular}} & & \multicolumn{1}{c}{\begin{tabular}[c]{@{}c@{}}Video \\ Games \end{tabular}} & & \multicolumn{1}{c}{Pornography}\\

\toprule

\multirow{3}{*}{\begin{tabular}[c]{@{}l@{}} OD-parse \end{tabular}} & Absolute  & 0.01 &&  -0.01  &&  0.07 \\
 & JS div.  & 0.01 && -0.01 &&   -0.01 \\
 & EMD  & 0.07 &&  0.01 &&  -0.01\\
 \toprule
 
\multirow{3}{*}{\begin{tabular}[c]{@{}l@{}} OD \end{tabular}} & Absolute  & \textbf{0.54} & &\textbf{0.56} 	&&  \textbf{0.41} \\
 & JS div.  & 0.07 &&  -0.01 &&  -0.02 \\
 & EMD  & 0.26 && -0.01 &&  0.01 \\

 \toprule
 
\multirow{3}{*}{\begin{tabular}[c]{@{}l@{}} OD (no \\ polarity \\ shifters) \end{tabular}} & Absolute  & 0.23 & &  0.08 &&  0.04 \\
 & JS div. & 0.09 &&  -0.01 &&  -0.02 \\
 & EMD &  0.10 &&  0.01 &&  -0.01 \\ 
 \toprule
\end{tabular}

}
\caption{We compare the quality of variants of Opinion Distance measures on opinion clustering task with ARI.} 
\label{tab:variants}
\end{table}

\begin{table*}[t]
\small
\centering

\resizebox{0.99\linewidth}{!}{
\begin{tabular}{@{}lr|rrrrrrr|rrrrrrr@{}}
\toprule
Topic Name & Size &  TF-IDF  &  WMD   & Sent2vec &  Doc2vec &  BERT   & $OD$-w2v  & $OD$-d2v  &   TF-IDF &  WMD & Sent2vec & Doc2vec  &  BERT & $OD$-w2v & $OD$-d2v \\

 &    &  ARI  &  ARI   &   ARI &   ARI &  ARI  &  ARI &  ARI  &   $Sil.$ &   $Sil.$ &   $Sil.$ &  $Sil.$ &  $Sil.$ &  $Sil.$ & $Sil.$ \\

\midrule

Affirmative Action    & 81 & -0.07 & -0.02 & 0.03 & -0.01 & -0.02 & \textbf{0.14} & {\it \underline{0.02}} & 0.01 & 0.01 & -0.01 & -0.02 & -0.04 & \textbf{0.06} & {\it \underline{0.01}} \\
 Atheism    & 116 & \textbf{0.19} & 0.07 & 0.00 & 0.03 & -0.01 & 0.11 & {\it \underline{0.16}}  & 0.02 & 0.01 & 0.02 & 0.01 & 0.01 & {\it \underline{0.05}} & \textbf{0.07} \\
 Austerity Measures    & 20 & {\it \underline{0.04}} & {\it \underline{0.04}} & -0.01 & -0.05 & 0.04 & \textbf{0.21} & -0.01 & 0.06 & 0.07 & 0.05 & -0.03 & 0.10 & \textbf{0.19} & 0.1 \\
 Democratization    & 76 & 0.02 & -0.01 & 0.00 & {\it \underline{0.09}} & -0.01 & \textbf{0.11} & 0.07 & 0.01 & 0.01 & 0.02 & 0.02 & 0.03 & \textbf{0.16} & {\it \underline{0.11}} \\
 Education Voucher Scheme    & 30 & \textbf{0.25} & 0.12 & 0.08 & -0.02 & 0.04 & 0.13 & {\it \underline{0.19}} & 0.01 & 0.01 & 0.01 & -0.01 & 0.02 & {\it \underline{0.38}} & \textbf{0.40} \\
 Gambling    & 60 & -0.06 & -0.01 & -0.02 & 0.04 & 0.09 & {\it \underline{0.35}} & \textbf{0.39} & 0.01 & 0.02 & 0.03 & 0.01 & 0.09 & \textbf{0.30} & {\it \underline{0.22}} \\
 Housing    & 30 & 0.01 & -0.01 & -0.01 & -0.02 & 0.08 & \textbf{0.27} & 0.01 & 0.02 & 0.03 & 0.03 & 0.01 & 0.11 & \textbf{0.13} & {\it \underline{0.13}} \\
 Hydroelectric Dams    & 110 & \textbf{0.47} &  {\it \underline{0.45}} & {\it \underline{0.45}} & -0.01 & 0.38 & 0.35 & 0.14 & 0.04 & 0.08 & 0.12 & 0.01 & 0.19 & \textbf{0.26} & {\it \underline{0.09}} \\
 Intellectual Property     & 66 & 0.01 & 0.01 & 0.00 & 0.03 & 0.03 & {\it \underline{0.05}} & \textbf{0.14} & 0.01 & {\it \underline{0.04}} & 0.03 & 0.01 & 0.03 & {\it \underline{0.04}} & \textbf{0.12} \\
 Keystone pipeline    & 18 & 0.01 & 0.01 & 0.00 & -0.13 & \textbf{0.07} & -0.01 & \textbf{0.07} & -0.01 & -0.03 & -0.03 & -0.07 & 0.03  & \textbf{0.05} & {\it \underline{0.02}} \\
 Monarchy    & 61 & -0.04 & 0.01 & 0.00 & 0.03 & -0.02 & \textbf{0.15} & \textbf{0.15} & 0.01 & 0.02 & 0.02 & 0.01 & 0.01 & \textbf{0.11} & {\it \underline{0.09}} \\
 National Service    & 33 & 0.14 & -0.03 & -0.01 & 0.02 & 0.01 & {\it \underline{0.31}} & \textbf{0.39} & 0.02 & 0.04 & 0.02 & 0.01 & 0.02 &  \textbf{0.25} & \textbf{0.25} \\
 One-child policy China    & 67 & -0.05 & 0.01 & \textbf{0.11} & -0.02 & 0.02 & \textbf{0.11} & 0.01 & 0.01 & 0.02 & {\it \underline{0.04}} & -0.01 & 0.03 & \textbf{0.07} & -0.02 \\
 Open-source Software    & 48 & -0.02 & -0.01 & {\it \underline{0.05}} & 0.01 & 0.12 & \textbf{0.09} & -0.02 & 0.01 & -0.01 & 0.00 & -0.02 & 0.03 & \textbf{0.18} & 0.01 \\
 Pornography   & 52 & -0.02 & 0.01 & 0.01 & -0.02 & -0.01 & \textbf{0.41} & \textbf{0.41} & 0.01 & 0.01 & 0.02 & -0.01 & 0.03 & \textbf{0.47} & {\it \underline{0.41}} \\
 Seanad Abolition     & 25 & 0.23 & 0.09 & -0.01 & -0.01 & 0.03 & {\it \underline{0.32}} & \textbf{0.54} & 0.02 & 0.01 & -0.01 & -0.03 & -0.04 & {\it \underline{0.15}} & \textbf{0.31} \\
 Trades Unions    & 19 & {\it \underline{0.44}} & {\it \underline{0.44}} & \textbf{0.60} & -0.05 & 0.44 &  {\it \underline{0.44}} & 0.29 & 0.1 & 0.17 & 0.21 & 0.01 & 0.26 & \textbf{0.48} & {\it \underline{0.32}} \\
 Video Games    & 72 & -0.01 & 0.01 & 0.12 & 0.01 & 0.08 & {\it \underline{0.40}} & \textbf{0.56} & 0.01 & 0.01 & 0.06 & 0.01 & 0.05 &  {\it \underline{0.32}} & \textbf{0.42} \\
 \hline
 Average & 54.67 & 0.09 & 0.07 & 0.08 & 0.01 & 0.08 & \textbf{0.22} & {\it \underline{0.20}} & 0.02 & 0.03 & 0.04 & -0.01 & 0.05 & \textbf{0.20} &{\it \underline{0.17}} \\
\bottomrule
\end{tabular}
}
\caption{Performance comparison of the distance measures on all 18 datasets. The semantic distance in opinion distance (OD) measure is computed via cosine distance over either Word2vec (OD-w2v with semantic distance threshold 0.6) or Doc2vec (OD-d2v with distance threshold 0.3) embeddings. $Sil.$ refers to Silhouette Coefficient. The second  best result is italicized and underlined. The ARI and Silhouette coefficients scores of both {\em OD} methods ({\em OD-d2v} and {\em OD-w2v}) are statistically significant (paired t-test) with respect to baselines at significance level 0.005. }
\label{table:all_topics}
\end{table*}

\subsection{Opinion Distance Drilldown}
\label{exp:all_topics}
In this section, we present a detailed drilldown of our proposed opinion distance measure.
We consider the following two variants of our measure:
{\em OD-d2v} and {\em OD-w2v} where the semantic distance is computed as cosine distance over Doc2vec embeddings (pre-trained on Wikipedia) and Word2vec embeddings (pre-trained on Google) respectively. The semantic threshold for OD-d2v is set at $0.3$ while for OD-w2v is set at $0.6$. In both the variants, we use the sentiment polarity shifters, a complete list of which are presented in Table~\ref{polarity_shifters}. 
We evaluate our distance measures in the unsupervised setting, specifically, evaluating the clustering quality using the Adjusted Rand Index (ARI) and Silhouette coefficient. 
We benchmark against the following baselines: WMD (which relies on word2vec embeddings), Doc2vec and TF-IDF. The results are shown in Table~\ref{table:all_topics}. The ARI and Silhouette coefficients scores of both {\em OD} methods ({\em OD-d2v} and {\em OD-w2v}) are statistically significant (paired t-test) with respect to baselines at significance level 0.005. We observe the following trends:
\begin{enumerate}

    \item Opinion distance methods generally outperform the competition on both ARI and Silhouette coefficient. We observe that given a topic, the opinion distance measure is able to separate pro stance opinions from the con stance opinions. We also find that the other baselines generally perform worse as both pro and con stance opinions have high text similarity in many of these datasets. This is reflected in the average ARI and average Silhouette coefficients of the baseline distance measures.

\item On a few datasets, there are a few discriminating words between the different opinion clusters. For instance, in the topic ``Trade Union", the opinion subject ``collective bargaining" is contained in 60 \% of claims in the con stance of the topic while it is not present in the pro stance opinions on the topic. The pro stance opinions on this topic use the term ``Unions" as the opinion subject. Another example is ``Hydroelectric Dams" dataset, where 36 \% of pro stance opinions contain term ``hydro" while only 18\% of con stance opinions contain term ``hydro". On such datasets, TF-IDF and WMD perform relatively better in separating pro stance opinions from the con stance opinions. In the exceptional case of ``Hydroelectric Dams" dataset, the opinion distance $OD$ performs particularly bad compared to TF-IDF because of some errors with disambiguation.

\end{enumerate}

\subsection{Variants of opinion distance}

Next, we compare the different variants of opinion distance. As mentioned earlier, the polarity of opinion subject $pol(S_{i})$ can be represented in the form of absolute value or polarity distribution.  The results of different variants are shown in Table~\ref{tab:variants}. We make the following observations.

 \textbf{$OD$ significantly outperforms $OD$-$parse$:} We observe that compared to $OD$-$parse$, $OD$ is much more accurate. On the three datasets, 
 $OD$ achieves an average weighted F1 score of 0.54, 0.56 and 0.41 respectively compared to the scores of 0.01, -0.01 and 0.07 by $OD$-$parse$. This is largely because of the errors in dependency parsing and the resultant poor opinion representation.

 \textbf{Representing polarities in form of distribution is inefficient:} We find that irrespective of the chosen variant of opinion distance, if we represent $pol(S_{i})$ in the form of distribution, we do not find good opinion clusters. The $D$ value of opinion distance calculated using polarity distribution is also low. Recall that Jensen-Shannon divergence is the distance measure used in ~\cite{FSSY12}. Table~\ref{tab:variants} shows that $OD$ is significantly better than Jensen-Shannon divergence on all the three datasets.

 \textbf{Sentiment polarity shifters have a high impact on clustering performance of opinion distance: } We find that not utilizing the sentiment polarity shifters, especially in case of datasets ``Video games" and ``Pornography" hurts the {\it Opinion Representation} phase, and thereby leads to incorrect computation of opinion distance. This is evident from the significant drop in ARI score from $OD$ to $OD$ (no polarity shifters) since the only change in those variants is of sentiment polarity shifters.

\section{Case study}

\begin{table}[t]
\small

\begin{tabularx}{\linewidth}{L} 
\textbf{Title: Brexit: Second Commons defeat for Theresa May in 24 hours} \\ \hline
{\it $O_1$}: ``What's most detestable is the way people on here protest about how Brexit is an almost divine right of the people which should be delivered as though a birthright or destiny. It's not. It's really not. It was a snapshot of public opinion at a particular time point which was influenced by a lot of spin and subterfuge. Now is a different time. Time for this horror to end." \\ \hline
\textbf{Below opinions $O_2$ and $O_3$ are similar to {\it $O_1$}} \\
{\it $O_2$} : ``Not all  leavers are thugs, i know because i voted leave and i'm not a thug. But there's a lot of leavers that have their head in the sand and can't admit that brexit is so complex that they and i didn't really understand the consequences. I'm not one of them leavers either as i've listened to the facts now and changed my mind. New factual vote please." \\ \hline
{\it $O_3$}: ``STOP BREXIT,  SAVE BRITAIN! They should jst go straight on to \#revokeA50. It would save lots of wasted time, money and even the economy. This is now crunch time for Brexit. It looks like the doors have been closed for a No Deal exit and Theresa May's plan will most likely be voted down. The only remaining option therefore is to call the whole thing off and let us get back to normal. " \\ \hline
\end{tabularx}
\caption{Similar comments identified with the help of the proposed OD measure on~\cite{bbcBrexit1}.}
\label{case_comments}
\end{table}

\begin{table}[h!t]
\small


\begin{tabularx}{\linewidth}{L} 
\textbf{Title: Theresa May urges Jeremy Corbyn: Let's talk Brexit} \\ \hline
{\it $O_1$}: ``The last thing Comrade Corbyn wants is cross-party consensus on Brexit. That would get in the way of his desperate plan to force a general election. It is now obvious to everyone (I hope) that this horrible man will say and do anything to get into power. He belongs in a glass case in a political museum, spouting his far-left communist trash, a bit like one of those old laughing sailor machines."\\ \hline
{\it $O_2$}: ``The longer this goes on, the less it has to do with the EU. Brexit has become a conflict between those who value truth and individual rights above all else, and those for whom being in a majority (real, ``silent" or based on a flawed referendum) is their identity. That's why the latter are so perpetually angry: they got what they wanted in 2016 but will lose it if facts win over fantasy in the end."\\ \hline
\end{tabularx}
\caption{Two comments expressing opposite opinions; OD correctly identifies them as dissimilar while Doc2vec considers them very similar.
}
\label{case_comments_dis}
\end{table}

Here we describe a case study for navigating opinions expressed in the form of comments on a BBC news article ``Brexit: Second Commons defeat for Theresa May in 24 hours"~\cite{bbcBrexit1}.
Table~\ref{case_comments} shows an example of comments that were correctly identified to be similar by OD, but reported highly dissimilar by TF-IDF and WMD. These comments are clearly against Brexit and want it to either be called off or be subjected to another referendum.
While in the above example the Doc2vec distance was also smaller, it fails to capture the dissimilarity between comments that are very different. For instance, Table~\ref{case_comments_dis} presents one comment that is highly critical of Labour leader Jeremy Corbyn, while the second comment opines poorly on the ruling conservative party; Doc2vec identifies them as similar while OD correctly identifies them as highly dissimilar.



\section{Conclusion}
Automated tools for better understanding of opinions would lead us to a new era of digital democracy and improved decision making. In this work, we proposed an opinion distance measure for quantifying the distance between opinions. Based on the observation that similar opinions express similar sentiment polarity on discussed subjects, we show that our proposed measure significantly outperforms existing approaches in both unsupervised and supervised experimental setups. 


 

\bibliographystyle{aaai}
\bibliography{sample-bibliography}

\newpage
\section{Appendix}


\subsection{Noun-phrase Representation: SemRegex Rules } %

The parse-tree variant of opinion distance $OD-parse$ relies on the dependency parse tree to extract opinion subject and opinion expression terms. In addition to the basic dependencies, enhanced dependencies and Named entity extraction functionalities present in Stanford CoreNLP, we also lever SemRegex rules. The list of rules with Stanford CoreNLP’s SemRegrex pattern matching system is presented in Table \ref{tab:semregex}. The $OpSubject$ refers to opinion subject while $OpExp$\# refers to the opinion expression. 

\begin{table}[!ht]
    \centering
    \def\arraystretch{1.4}
   \resizebox{1.0\linewidth}{!}{
   \small{
    \begin{tabular}{c|l}
    \toprule
         & \textbf{SemRegex Rules} \\
         \midrule
1 & $ \{\}=OpExp1 >/nmod:in/ \{\}=OpSubject$  \\ 
2 & $ \{\}=OpExp1 >/nmod:to/ \{\}=OpSubject$ \\
3 & $ \{\}=OpExp1 </nmod:of/ \{\}=OpSubject$ \\
4 & $ \{\}=OpExp1 >/nmod:by/ \{\}=OpSubject$ \\
5 & $ \{\}=OpExp1 >/nmod:after/ \{\}=OpSubject$ \\
6 & $ \{\}=OpExp1 >/nmod:without/ \{\}=OpSubject$ \\
7 & $ \{\}=OpExp1 >nsubj \{\}=OpSubject > $\\ 
& $ dobj \{\}=OpExp2$ \\
8 & $ \{\}=OpExp1 >nsubj \{\}=OpSubject < $\\ 
& $ csubj \{\}=OpExp2$ \\
9 & $ \{\}=OpExp1 >nsubj \{\}=OpSubject > $\\ 
& $ /compound.*/ \{\}=OpExp2$ \\
10 & $ \{\}=OpExp1 </advcl:as/ \{\}=OpSubject$ \\
11 & $ \{tag:/VB.*/\} >nsubj \{\}=OpSubject  $\\ 
& $[ >acomp \{\}=OpExp1 | >xcomp \{\}=OpExp2]$ \\
12 & $ \{tag:/VB.*/\}=OpExp2 >advmod $\\ 
& $\{\}=OpExp1 [< \{\}=OpSubject | > \{\}=OpSubject]$ \\
13 & $ \{tag:/NN.*/\}  >nsubj \{\}=OpSubject $\\ 
& $ >amod \{\}=OpExp1$ \\
14 & $ \{tag:/NN.*/\}  >nsubj \{\}=OpSubject $\\ 
& $ >/nmod.*/ (\{\}=OpExp1 >amod \{\}=OpExp2$ \\
         \bottomrule
    \end{tabular}
    }
  }
    \caption{SemRegex Rules}
    \label{tab:semregex}
\end{table}

\end{document}